\documentclass[11pt]{article}
\usepackage[utf8]{inputenc}
\usepackage[T1]{fontenc}
\usepackage[margin=1in]{geometry}
\usepackage{amsmath,amssymb}
\usepackage{booktabs}
\usepackage{array}
\usepackage{tabularx}
\usepackage{enumitem}
\usepackage{microtype}
\usepackage[hidelinks]{hyperref}
\usepackage{xurl}
\newcolumntype{L}{>{\raggedright\arraybackslash}X}
\title{Skill Blocks: How Should an Agent Load Its Skill? A Caching-Correct Comparison of Pre-load, On-Demand Tool-Loading, Progressive Disclosure, and Hybrid}
\author{Hironobu Nakasuji\\Microsoft}
\date{}
\begin{document}
\maketitle
\begin{abstract}
Agent skills are instruction documents that give an AI agent specialized
knowledge. A common implementation places the entire skill in every request,
even when most of it is irrelevant. This can waste tokens, especially for
large skills and multi-turn tasks. We ask a practical systems question: given
a fixed skill, when should each part be shown to the model?

We compare four loading methods. \textbf{Full} always includes the complete
skill. \textbf{Skill Block} begins with a small core and lets the model request
additional sections through a tool. \textbf{Reference} begins with a catalog
and then supplies a selected reference section. \textbf{Hybrid} provides short
summaries first and loads more detail only when needed. The same procedures
remain available across the main comparisons, although ALFWorld also differs
in how guidance is exposed. We additionally report a separately run
SpreadsheetBench treatment, \texttt{original8}, that permanently removes six
sections and therefore tests pruning rather than conditional loading.

We evaluate these methods on five paired benchmarks. For single-turn tasks, we
compare the input tokens used for each item. For multi-turn tasks, many tokens
are reported as cache reads, so we also use an ``effective input'' measure that
counts new tokens fully and cache-read tokens at 10\% of their raw amount.
There is no universal winner. On SearchQA (n=1,400), hybrid reduces input by
27.4\%, while heavily encouraged Skill Block loading increases it by 48.4\%.
On SpreadsheetBench, hybrid, Skill Block, and reference reduce input by
39.8\%, 35.6\%, and 31.7\% on 276 token-complete cases; the non-content-parity
\texttt{original8} treatment reduces it by 55.5\%. On ALFWorld (n=42), the
procedures are short and repeatedly needed, so Skill Block's 12.6\% input
reduction becomes small after also accounting for its longer outputs. Larger,
more compressible skills show clearer gains. On the primary ScienceWorld set
(n=84), Skill Block and hybrid reduce effective input by 62.5\% and 52.8\%;
on SynthProc (n=40), they reduce it by 73.0\% and 66.6\%. Paired outcome tests
detect no quality difference in the principal comparisons, but they do not
prove equivalence or rule out regression. Additional runs configured with a
second model retain the large-skill pattern, with limitations from
same-provider testing, unavailable provider-confirmed model identity, and a
SynthProc refusal confound. Overall, conditional loading is most useful when a
skill contains substantial material that is not needed on every turn; for
small or frequently used skills, loading overhead can erase the savings.

\end{abstract}
\section*{1. Introduction}
Skills are an increasingly common interface for specializing agent behavior without changing model weights. Agent Skills and related \texttt{SKILL.\allowbreak{}md} conventions package procedures, constraints, examples, and tool-use guidance into files that a model reads at inference time [Anthropic 2025a; Anthropic 2025b]. The Model Context Protocol (MCP) provides a standardized context-delivery interface [Anthropic 2024]. At the same time, methods such as SkillOpt optimize the content of a skill, alongside automated prompt and pipeline optimization methods including APE, DSPy, TextGrad, and GEPA [Yang et al. 2026; Zhou et al. 2023; Khattab et al. 2024; Yuksekgonul et al. 2024; Agrawal et al. 2025].

The deployment cost of this pattern is easy to overlook. A monolithic skill is normally concatenated to every prompt, even if a particular request needs only one of its procedures. The result is a breadth--cost coupling: adding coverage makes every call longer. Prompt length can also affect how models use information in context [Liu et al. 2024]. A direct alternative is to compress the prompt [Jiang et al. 2023; Jiang et al. 2024]. We study a different, complementary choice: retain the source content but select which blocks enter context.

Our question is therefore \textbf{how to load an already-good skill}, not how to write a better one. We convert a skill into a lossless block substrate and compare four packagings. \texttt{full} sends all blocks. \texttt{Skill Block} keeps a small core in context and lets the model call a tool for optional blocks. \texttt{reference} uses progressive disclosure: a compact catalog plus model-requested reference files. \texttt{hybrid} sends short stubs for all blocks and fetches complete blocks only when a stub is insufficient. In the four primary mechanisms, the same content is available and full rendering recovers the monolithic skill. The prompt scaffolding necessarily differs---tools, references, and stubs require distinct instructions---so the estimand is packaging \textbf{plus required scaffolding}, not a byte-identical prompt intervention.

This framing resembles conditional computation in mixture-of-experts systems [Shazeer et al. 2017; Fedus et al. 2022], but gating here selects natural-language prompt blocks rather than learned parameter experts. It also differs from external retrieval: the candidate corpus is the skill itself and the selected text is copied verbatim. Whether selection pays depends on how much optional text is unused, how often a mechanism incurs a loading round-trip, and how much fixed context is resent. Consequently, neither ``on demand'' nor any named mechanism is automatically cheaper.

A second complication is cache accounting. Stateful agents resubmit a growing conversation. Provider usage reports include cache-read tokens in \texttt{input\_\allowbreak{}tokens}; treating all of them as newly processed can materially misstate multi-turn cost. We therefore report raw input, new input, and an effective sensitivity that prices cache reads at a discount. This correction does not change single-turn comparisons, where raw input is the relevant primary metric.

We make four contributions. First, we give a content-preserving substrate and a four-mechanism comparison of skill loading. Second, we formalize the unused-mass versus overhead trade-off and apply caching-correct accounting. Third, we evaluate five regimes spanning skill size and turn structure: SearchQA [Dunn et al. 2017], SpreadsheetBench [Ma et al. 2024], ALFWorld [Shridhar et al. 2021], ScienceWorld [Wang et al. 2022], and a controlled synthetic probe, SynthProc. ALFWorld is grounded in ALFRED-style tasks [Shridhar et al. 2020] through TextWorld [Côté et al. 2018]; the interactive agents use ReAct-style reasoning and acting [Yao et al. 2023]. Fourth, we rerun the large-skill multi-turn cases with a second configured endpoint model, \texttt{gpt-\allowbreak{}5.6}, while retaining the scope limits of same-provider, configuration-level transfer.

The result is a regime map. Hybrid is the best content-parity mechanism in the two single-turn cells, but static pruning can still be cheaper when reduced coverage is acceptable. For large, compressible multi-turn skills, Skill Block and hybrid lead; for small, always-needed multi-turn procedures, the advantage is near parity under broader output accounting. These are paired within-benchmark findings, not a claim of equivalence, reliability, or a universal loading rule.

\section*{2. Related Work}
\textbf{Skills and optimization.} Agent Skills and MCP establish an operational setting for reusable instruction documents and context delivery [Anthropic 2025a; Anthropic 2025b; Anthropic 2024]. SkillOpt seeks better skill content through validation-gated edits [Yang et al. 2026]. APE, DSPy, TextGrad, and GEPA likewise optimize prompts or pipelines [Zhou et al. 2023; Khattab et al. 2024; Yuksekgonul et al. 2024; Agrawal et al. 2025]. Our intervention is orthogonal: with content fixed, it changes when blocks are supplied. It can thus follow content optimization rather than compete with it.

\textbf{Conditional context and compression.} Sparse MoE routes inputs to subsets of parameters [Shazeer et al. 2017; Fedus et al. 2022]. Skill loading routes inputs to subsets of text, using a model-mediated semantic choice rather than a trained gate. Prompt compressors reduce a prompt representation, usually lossily [Jiang et al. 2023; Jiang et al. 2024]; our selected blocks remain verbatim, and compression could be applied after selection. This distinction matters for interpreting results: we test availability-preserving packaging, not a new compressed skill.

\textbf{Agent evaluation.} We deliberately span context-dominated QA, spreadsheet manipulation, text embodiment, and an interactive scientific simulator [Dunn et al. 2017; Ma et al. 2024; Shridhar et al. 2021; Wang et al. 2022]. SynthProc complements these benchmarks with a controlled per-step knowledge gate; it is not a substitute for a natural task. The benchmark choices expose the two variables central to our analysis---how much skill text is optional and whether its always-on footprint recurs over turns.

\section*{3. Method}
\subsection*{3.1 Shared substrate and mechanisms}
A skill is partitioned into named blocks with a natural-language description, \texttt{default} flag, dependencies, priority, and tags. One or more core blocks are always available; optional blocks can depend on other blocks. Rendering all blocks reproduces the monolithic skill (byte-for-byte for block-authored skills and subject only to whitespace/heading normalization when converting an existing document). When a block is loaded, dependencies already in context are not duplicated. Thus the four principal arms retain the same available skill content.

\begin{center}\small
\begin{tabularx}{\linewidth}{LLLL}
\toprule
mechanism & always-on context & on-demand path & distinguishing cost \\
\midrule
\textbf{full} & rendered skill, all blocks & none & full optional mass \\
\textbf{Skill Block} & core + tool schema/catalog & \texttt{load\_\allowbreak{}skill\_\allowbreak{}block(name)} returns block verbatim & schema + load turn \\
\textbf{reference} & core + text catalog & model selects reference file, then answers & selection turn; no tool schema \\
\textbf{hybrid} & core + compact stubs + tool schema & full block only if stub is insufficient & stub; occasional load turn \\
\bottomrule
\end{tabularx}
\end{center}

\texttt{full} is the conventional pre-load baseline. Skill Block has the lightest text core but carries a function schema and can require a tool round-trip. Reference is a faithful progressive-disclosure implementation: it has no tool schema but requires a selection round before its answer. Hybrid supplies a short quick-reference view of each block; it can avoid a fetch in cases where that view is sufficient. The catalog and loading instructions are mechanism-required scaffolding, not held-constant content.

The exact hybrid tiering is benchmark-specific. SearchQA, SpreadsheetBench, and historical SynthProc use a compact stub followed by the complete block. Historical ALFWorld and ScienceWorld use the retained \texttt{truehybrid} implementation: a first on-demand call returns a lean hint and pointer, and a second can return the full reference. We report both as hybrid because they instantiate progressive two-stage disclosure, but do not treat their tool-call counts as the same intervention.

A static reduced skill is a different practical alternative. On SpreadsheetBench we also report historical \texttt{original8}, which permanently omits six blocks and registers no loader. Its eight blocks are not identical to the dynamic arms' exact \texttt{CORE\_\allowbreak{}BLOCKS}; it is not content parity, was run separately, and cannot establish that it covers unobserved deployment demand.

\subsection*{3.2 Cost and caching model}
Let \texttt{c} be always-on core tokens, \texttt{s\_\allowbreak{}i} optional block sizes, \texttt{p\_\allowbreak{}i} their load probabilities, \texttt{F} fixed harness/system overhead, \texttt{X} task context, \texttt{T} tool-schema tokens, and \texttt{r} the number of model turns in a loading interaction. An input-side approximation is

\begin{align*}
\mathbb{E}[\text{skill tokens}] &= c + \sum_i p_i s_i \\
\mathbb{E}[\text{input}_{\text{on-demand}}] &\approx r(F+c+T+X) + \sum_i p_i s_i \\
\text{input}_{\text{full}} &= F+c+\sum_i s_i+X \\
\text{net saving} &\approx \sum_i(1-p_i)s_i - \big[(r-1)(F+c+X)+rT\big]
\end{align*}

For reference, \texttt{T=0} but its selection normally makes \texttt{r=2}; for hybrid, stubs alter the first-tier term and a low fetch rate can leave $r\approx 1$. The expression is an accounting model, not a learned selector or a prospective performance guarantee. It states the break-even condition that unused optional mass must outweigh round-trip and schema overhead. In multi-turn settings, the persistent prompt footprint and schema are presented repeatedly, making the always-on footprint especially important.

Reported multi-turn \texttt{input\_\allowbreak{}tokens} include cache reads. We compute

\[
\begin{aligned}
\text{new input} &= \text{input tokens} - \text{cache read}, \\
\text{effective input} &= \text{new input} + d\cdot\text{cache read}, \qquad d=0.1.
\end{aligned}
\]

Raw input corresponds to \texttt{d=1} and is an upper bound on this input-cost sensitivity. We use raw input as the primary metric for single-turn experiments: each case is one benchmark interaction, although an arm may make multiple provider calls, and any batch cache reuse is a cross-case harness artifact rather than within-case cost. For multi-turn experiments, primary tables use median effective input. Generated tokens are not free, so we also report \texttt{total\_\allowbreak{}\ensuremath{\lambda} = base\_\allowbreak{}input + \ensuremath{\lambda} \ensuremath{\cdot} output\_\allowbreak{}tokens}, where \texttt{base\_\allowbreak{}input} is raw input for single-turn and effective input for multi-turn. $\lambda  = 1, 4, or 8$ is a provider-neutral sensitivity, not a GHCP billing assertion. Historical AgentBoard logs did not expose a separate reasoning-token field.

A deployment may estimate this expression and inline a mechanism when estimated overhead exceeds savings. This gate is \textbf{mechanism-specific}: the observed small SearchQA hybrid case has a near-zero fetch rate while pure Skill Block does not. It is a post-hoc engineering heuristic, not a validated predictor or universal size threshold. Guidance-section implementation and dependency rendering are detailed in Appendix A.

\section*{4. Experimental Setup}
\subsection*{4.1 Protocol, metrics, and disclosures}
Most primary experiments used a GitHub Copilot production endpoint configured with \texttt{GHCP\_\allowbreak{}LEAN\_\allowbreak{}SYSTEM=1} and reported as \texttt{gpt-\allowbreak{}5.5}. Because its system prompt and decoding settings are not public, the absolute scores apply to this endpoint and should not be read as universal model scores. The historical logs also do not independently confirm every model identifier returned by the provider. SearchQA, in particular, records only the endpoint's historical default configuration; it does not retain a provider-returned model identity or reasoning effort. Within each benchmark, however, every arm evaluates the same items on the same endpoint, so our claims are based on paired comparisons under a shared configuration. To check whether the findings transfer to a different configuration, we reran the complete SynthProc and ScienceWorld suites with \texttt{GHCP\_\allowbreak{}MODEL=gpt-\allowbreak{}5.6-\allowbreak{}terra} and left reasoning effort unset. Those logs likewise do not retain provider-returned model identifiers, and a separate medium-effort SynthProc run is included only as supplementary evidence.

When the harness retains token telemetry, we record input, cache-read, and output tokens. We also check that \texttt{new=input-\allowbreak{}cache\_\allowbreak{}read} and that \texttt{cache\_\allowbreak{}read\ensuremath{\leq}input}. Historical ScienceWorld, ALFWorld, and SynthProc results contain episode-level totals rather than complete call-by-call telemetry. Only the 24 targeted ScienceWorld replacement runs have been audited at the call level.

SearchQA reports both strict (``hard'') and lenient (``soft'') normalized answer matches. The retained usage totals cover 1,400/2,453/1,420/2,800 completed provider calls for full/Skill Block/hybrid/reference, respectively. They exclude usage from 3/5/1/17 failed sends. A non-destructive parser correction changes one retained Skill Block answer but does not change any token totals. Because most rows retain the extracted prediction rather than the complete raw response, this correction is not an exhaustive rescore of all raw responses.

SpreadsheetBench includes 280 attempted cases, 279 model-completed cases, 276 cases with complete token data in every arm, and 275 cases with clean strict-value outcomes. A time-normalization correction changes Skill Block case \texttt{49667} from hard 0 to 1. Here, ``cell accuracy'' is the fraction of target cells that match; for the usable one-workbook cases, the benchmark's soft score is the same as workbook pass. Three evaluator-failed cases have unrecoverable usage. The retained research tree also contains 15 earlier full-test sweep directories spanning 23 arm evaluations. Because the final split was reused extensively, all SpreadsheetBench inference is exploratory. ALFWorld has one failed adaptive provider attempt with unrecoverable usage. ALFWorld and SynthProc report whether the complete episode succeeds, whereas ScienceWorld reports both binary completion and AgentBoard progress, defined as the fraction of scored subgoals completed.

Token intervals are paired 95\% confidence intervals from 100,000 bootstrap resamples, and outcome comparisons use paired exact McNemar tests. Across the nine principal \texttt{gpt-\allowbreak{}5.5} accuracy comparisons, every Holm-adjusted p-value is 1.0 under the corrected ScienceWorld denominator. This inferential family is exploratory because SearchQA and SpreadsheetBench were evaluated iteratively on the final cases, and SpreadsheetBench block selection used final-split outcomes. The tests show no detected difference; they do not establish non-inferiority or equivalence. The three configured-\texttt{gpt-\allowbreak{}5.6} ScienceWorld outcome comparisons form a separate family, with Holm-adjusted p=.116.

The historical ScienceWorld run used a 25-action cap, a 600-second between-turn decision deadline, and an intended 780-second process backstop; enforcement was not a strict 600-second episode wall clock. For full/Skill Block/hybrid/reference, abort counts are 1/2/2/3, records reaching 25 actions are 31/25/28/36, and non-aborted records at or above 600 seconds are 0/0/1/1. The eight abort-flagged arm-items span six IDs and include four zero-turn failures with missing usage. The primary analysis therefore uses the common four-arm set after excluding every abort-affected ID (n=84). A stricter n=82 sensitivity also excludes both overrun-affected IDs from every arm. We later reran the six abort-affected IDs under all four arms, replacing 24 rows in a derived matched n=90 artifact. Those replacements are provider-observed as \texttt{gpt-\allowbreak{}5.5}, with configured effort unset and observed effort \texttt{medium}; because they were run later and only those 24 rows have call-level cost auditing, the repaired n=90 result is a temporally spliced sensitivity, not the primary estimate.

\begin{center}\small
\begin{tabularx}{\linewidth}{LrLL}
\toprule
benchmark & n & interaction / skill regime & primary input metric \\
\midrule
SearchQA & 1,400 & single-turn; $\approx$2K skill & raw \\
SpreadsheetBench & 276 / 275 & single-turn; token-complete / outcome-clean; execution feedback & raw \\
ALFWorld & 42 & multi-turn; $\approx$1K, small procedures always needed & effective \\
ScienceWorld & 84 & multi-turn; $\approx$6K, 16 procedures, one/episode; common non-aborted primary set & effective \\
SynthProc & 40 & multi-turn; $\approx$10K, 45-operation library (44 exercised), different block/step & effective \\
\bottomrule
\end{tabularx}
\end{center}

\begin{center}\small
\begin{tabularx}{\linewidth}{LLrL}
\toprule
benchmark & blocks / tier-1 construction & full skill & key packaging detail \\
\midrule
SearchQA & overview + 5 optional & $\approx$1,966 & hybrid stub $\approx$385 tokens \\
SpreadsheetBench & \texttt{CORE\_\allowbreak{}BLOCKS} + 6 specialized optional & $\approx$8,116 & dynamic core $\approx$3,513; static \texttt{original8} $\approx$3,195 \\
ALFWorld & overview + 6 procedures & $\approx$1,087 & procedure walkthroughs average $\approx$99 tokens \\
ScienceWorld & overview + 16 procedures & $\approx$6,097 & walkthroughs average $\approx$289 tokens \\
SynthProc & overview + 45 operations & $\approx$9,728 & \texttt{o200k\_base}; hints plus key-bearing detail \\
\bottomrule
\end{tabularx}
\end{center}

Each mechanism has access to the same source content, but the prompts cannot be identical because the mechanisms present and retrieve that content differently. Some comparisons also include differences in retrieval guidance, which we disclose below.

In SearchQA, the frozen Skill Block arm uses an aggressive instruction to ``load before answering'' so that its overhead is visible. In preliminary runs on the first 200 items, alternative instructions caused fewer loads. Those pilots did not cover the full test set under controlled conditions, however, so they are not treated as formal ablations.

ALFWorld explicitly compares both the retrieval mechanism and its guidance. Its six quick hints contain 701 characters, or about 176 proxy tokens, and the reference arm receives all six. In the retained \texttt{truehybrid} results, 38 of 42 episodes record one tool call, two record two calls, and two record none. The two-call episodes are consistent with loading both a hint and a reference, but the historical logs do not identify the stage of each call.

ScienceWorld also uses \texttt{truehybrid}, and its logs retain aggregate counters for each retrieval stage. Across the primary 84 episodes, hybrid makes 81 hint loads and 54 reference loads. The numbers of episodes with 0/1/2/3/4/5 hint calls are 16/56/6/4/1/1, while the numbers with 0/1/2 reference calls are 31/52/1.

SpreadsheetBench evaluates selection on the full split, but execution feedback can recover from a poor selection. Similar final outcomes therefore do not directly show that routing decisions were equally accurate. SynthProc is a synthetic benchmark designed to isolate token mechanics rather than task reliability: every arm in the historically reported \texttt{gpt-\allowbreak{}5.5} run succeeds on all 40 tasks. These tasks exercise 44 of the 45 operations---only \texttt{op-fetch-records} (a keyed synthetic operation that retrieves records by category) is unused---and each task contains between zero and four keyless steps. Across all tasks, the aggregate split is 80 keyed and 80 keyless steps.

\subsection*{4.2 Evaluation interpretation}
Comparisons are paired on identical items. Stateful turn usage is aggregated per episode; medians are primary because long trajectories skew means, which remain a robustness check. Raw input is primary for SearchQA/SpreadsheetBench; effective input is primary for stateful ALFWorld/ScienceWorld/SynthProc. Token intervals concern the logged accounting quantity, not a universal cloud bill.

Outcome metrics remain native to each benchmark: SearchQA answer match, SpreadsheetBench workbook pass plus separately labeled cell accuracy, ALFWorld/SynthProc completion, and ScienceWorld binary completion plus progress. A non-significant outcome test is ``no detected difference,'' not equivalence. Static \texttt{original8} changes content availability and is outside the lossless four-arm comparison. Detailed metric semantics and interpretation are in Appendices A, C, and E.

\section*{5. Results}
All comparisons are paired by item. SearchQA and SpreadsheetBench values below are \textbf{mean raw input}; multi-turn values are \textbf{median effective input}. Full per-arm tables, paired intervals, and accounting are in Appendices B--D.

\subsection*{5.1 Consolidated \texttt{gpt-\allowbreak{}5.5} results}
\begin{center}\scriptsize
\begin{tabularx}{\linewidth}{Lrrrrr}
\toprule
benchmark (input n / quality n) & full & Skill Block & hybrid & reference & static \\
\midrule
SearchQA raw / hard (1,400 / 1,400) &
\shortstack{5,535\\.841} & \shortstack{+48.4\%\\.841} &
\shortstack{\textbf{$-$27.4\%}\\.845} & \shortstack{+10.4\%\\.849} & --- \\
SpreadsheetBench raw / pass (276 / 275) &
\shortstack{12,461\\.800} & \shortstack{$-$35.6\%\\.811} &
\shortstack{\textbf{$-$39.8\%}\\.807} & \shortstack{$-$31.7\%\\.807} &
\shortstack{\textbf{$-$55.5\%}\\.811} \\
ALFWorld effective / success (42 / 42) &
\shortstack{6,368\\42/42} & \shortstack{\textbf{$-$12.6\%}\\42/42} &
\shortstack{$-$3.2\%\\42/42} & \shortstack{+24.2\%\\42/42} & --- \\
ScienceWorld effective / success; progress (84 / 84) &
\shortstack{26,810\\57/84; .901} & \shortstack{\textbf{$-$62.5\%}\\62/84; .916} &
\shortstack{$-$52.8\%\\58/84; .902} & \shortstack{$-$23.0\%\\50/84; .833} & --- \\
SynthProc effective / success (40 / 40) &
\shortstack{21,617\\40/40} & \shortstack{\textbf{$-$73.0\%}\\40/40} &
\shortstack{$-$66.6\%\\40/40} & \shortstack{$-$14.8\%\\40/40} & --- \\
\bottomrule
\end{tabularx}
\end{center}

Each cell shows input (full level or delta) above the benchmark-native quality
outcome. These are observed levels, not non-inferiority: paired outcome tests
detect no difference in the principal family, but do not prove that quality
cannot regress.

\textbf{Single turn.} SearchQA is a small-skill, context-dominated case. Hybrid replaces the full skill with a stub and fetches only 0.02 blocks/query (1.4\% load anything), yielding 4,016 versus 5,535 raw tokens ($-$27.4\%; paired CI [$-$1,556, $-$1,479]). Its hard accuracy is .845. Reference costs 6,112 (+10.4\%), and aggressively nudged, ungated Skill Block costs 8,216 (+48.4\%; 2.05 loads/query). After the one-row parser correction, the hard range is .841--.849, so token differences---not outcome ranking---carry the result. At $\lambda$=4, hybrid remains $-$25.7\%, whereas reference is +22.2\% and Skill Block +69.7\%. The small-skill exception is important: hybrid's low fetch rate makes it viable; it does not validate pure tool loading at this scale.

SpreadsheetBench has a larger skill. On the 276 common token-complete cases, all content-parity on-demand arms beat full: hybrid 7,506 ($-$39.8\%), Skill Block 8,028 ($-$35.6\%), and reference 8,517 ($-$31.7\%) versus 12,461. On the 275 strict value-outcome-clean cases, hard workbook pass is .800 full, .811 Skill Block, and .807 hybrid/reference; no paired outcome comparison detects a difference. Specialized blocks are selected in 97/276 Skill Block cases, 47/276 hybrid cases, and 254/276 reference cases. The separately timed static \texttt{original8} treatment is 5,547 ($-$55.5\%), hard .811, and cell accuracy .903; $\lambda$=4 gives $-$38.4\% original8, $-$30.9\% hybrid, $-$26.5\% Skill Block, and $-$19.8\% reference. Because \texttt{original8} differs from exact \texttt{CORE\_\allowbreak{}BLOCKS}, it is an exploratory pruning comparison rather than a clean static-core causal control.

\textbf{Multi-turn.} ALFWorld is the boundary regime. Small procedures are always needed, leaving little optional mass to remove. Skill Block is 5,568.65 effective tokens ($-$12.55\%, aligned 95\% CI [$-$15.93\%, $-$7.45\%]), hybrid 6,165.75 ($-$3.18\%, [$-$9.50\%, +1.29\%]), and reference 7,906.85 (+24.17\%, [+19.41\%, +31.05\%]) versus 6,368 full; every arm completes 42/42 episodes. Hybrid therefore has no statistically established input saving under the aligned estimand. At $\lambda$=4, Skill Block is $-$4.5\%, hybrid +9.2\%, and reference +21.3\%; at $\lambda$=8, Skill Block is $-$0.6\%.

The two large, compressible multi-turn cells differ. On real ScienceWorld, one of 16 procedures is generally relevant per episode. In the primary common non-aborted set, Skill Block reduces effective input to 10,061 ($-$62.47\%, aligned 95\% CI [$-$67.76\%, $-$57.44\%]), hybrid to 12,654 ($-$52.80\%, [$-$64.74\%, $-$37.66\%]), and reference to 20,644 ($-$23.00\%, [$-$40.21\%, +4.49\%]), from 26,810 full. Success/progress are 62/84/.916 for Skill Block, 58/84/.902 for hybrid, 50/84/.833 for reference, and 57/84/.901 for full. Exact McNemar p-values are .180, 1.000, and .118; no outcome comparison supports superiority or equivalence. The repaired matched n=90 sensitivity gives reductions of 62.85\%, 50.84\%, and 25.38\%, but combines six later reruns with 84 historical IDs; its reference interval only narrowly excludes zero and is not robust across denominator or output-weight sensitivities. On SynthProc, a different block can be needed at each of four steps: Skill Block is 5,837 ($-$73.0\%), hybrid 7,221 ($-$66.6\%), and reference 18,428 ($-$14.8\%) versus 21,617 full, with 40/40 completion in all arms. Thus SynthProc is a controlled token result, while ScienceWorld is the load-bearing real-task evidence.

\subsection*{5.2 Robustness and accounting}
The principal multi-turn metric matters because cache reads make up 74--94\% of raw input: 87--92\% in ALFWorld, 93--94\% in ScienceWorld, and 74--90\% in SynthProc. Raw counts can therefore misrank an arm. For example, SynthProc reference has raw input 70,528, above full's 65,380, but effective input 18,428, 14.8\% below full. This is an accounting inversion, not proof that cache attribution is known: logs are consistent with repeated-prefix caching, but do not isolate whether schema tokens, conversation tokens, or both received cache treatment.

Medians are primary because episode costs are skewed. Means preserve the qualitative ordering: ALFWorld Skill Block/hybrid/reference mean deltas are $-$13.6\%/$-$2.4\%/+18.5\%; ScienceWorld $-$53.3\%/$-$37.7\%/$-$20.2\%; SynthProc $-$73.1\%/$-$67.0\%/$-$14.9\%. The ScienceWorld median $-$62.5\% should therefore not be read as its only summary. Output sensitivity also preserves the large-skill result more clearly than the ALFWorld boundary: at $\lambda$=4, Skill Block/hybrid/reference are $-$46.8\%/$-$24.5\%/$-$2.7\% on ScienceWorld and $-$53.7\%/$-$48.0\%/$-$10.2\% on SynthProc. At $\lambda$=8, ScienceWorld reference reverses to +12.5\%, whereas Skill Block and hybrid remain below full.

The cache discount is a caveat rather than a universal price. Large-skill multi-turn advantages remain on raw input and hence do not rely on choosing \texttt{d=.1}. ALFWorld does: its paired-median crossover is $d\approx .327$, so its Skill Block lead holds at the stated reference discount but not at all discounts. We count new tokens at full rate and do not add a cache-write premium. This is conservative for winning on-demand arms in the available ALFWorld write telemetry (Skill Block 2,619 and hybrid 2,960 median cache-write tokens versus full 3,884; reference 4,544). Detailed \texttt{d}, $\lambda$, raw/new/effective, write, interval, and Holm tables appear in Appendix C.

\subsection*{5.3 Configured model transfer}
The within-run paired default-effort SynthProc check configured as \texttt{gpt-\allowbreak{}5.6} (n=40) gives marginal effective medians of 14,053 full, 4,725 Skill Block ($-$66.4\%), 4,731 hybrid ($-$66.3\%), and 11,760 reference ($-$16.3\%); success is 37/40, 40/40, 40/40, and 39/40, respectively. The six-token marginal gap does not justify a stable ordering across configurations. The supplementary medium-effort SynthProc result is hybrid $-$68.4\%, Skill Block $-$66.4\%, and reference $-$8.9\%, again a configured-endpoint check. Key-specific non-call actions containing ``secret,'' ``credential,'' or ``keyed'' affect 13/40 default full episodes (10 still successful), 2/40 default reference episodes (1 successful), 11/40 medium full episodes (6 successful), and 4/40 medium reference episodes (3 successful); this classifier flags no Skill Block/hybrid episode. A broader inability diagnostic gives default full 16/40 (13 successful), default hybrid 1/40 (successful), and medium full 12/40 (7 successful). The confound therefore changes token cost as well as failures and prevents a reliability interpretation or clean loading-effect attribution.

The complete real ScienceWorld rerun configured as \texttt{gpt-\allowbreak{}5.6} (n=90) is more informative. Effective input is 28,459 full, 8,869 hybrid ($-$68.8\%), 9,173 Skill Block ($-$67.8\%), and 11,812 reference ($-$58.5\%). Success/progress are 52/90/.859, 60/90/.904, 58/90/.875, and 60/90/.896 for full/hybrid/Skill Block/reference. Paired median effective deltas are $-$13,916 for Skill Block (CI [$-$16,014, $-$11,900], cheaper 89/90), $-$13,683 for hybrid (CI [$-$16,645, $-$11,020], 89/90), and $-$11,461 for reference (CI [$-$13,241, $-$8,071], 88/90). At $\lambda$=4 the reductions remain $-$63.9\%, $-$63.4\%, and $-$54.6\%; corrected outcome tests have Holm-adjusted p=.116. Together, the two configured runs support same-provider configuration-transfer evidence for the large, compressible on-demand regime, not provider-attested model isolation, cross-vendor replication, or model independence.

\subsection*{5.4 Regime map and decision rule}
The cost model organizes the observed cells after the fact. For single-turn use, first test whether a fixed reduced skill preserves required coverage. The SpreadsheetBench static control is cheapest on its measured split, but it is not content parity. If all optional content must remain available, hybrid is the best observed content-parity choice in both single-turn cells: it preserves a compact view while avoiding most fetches. Pure Skill Block should be inlined when its expected schema and round-trip overhead exceeds optional mass; SearchQA is the motivating ungated failure.

For multi-turn use, assess whether the skill is large and compressible. When substantial text is unused across repeated turns, the lean on-demand pair is favored: reported-\texttt{gpt-\allowbreak{}5.5} ScienceWorld and SynthProc place Skill Block first, while the configured-\texttt{gpt-\allowbreak{}5.6} runs make hybrid marginally first on ScienceWorld and effectively tie the pair on SynthProc. When procedures are small and always needed, as in ALFWorld, treat any apparent input win as near parity after outputs and cache-discount sensitivity. Reference never leads in these five cells, but that empirical pattern is not a dominance theorem.

\begin{center}\small
\begin{tabularx}{\linewidth}{LLL}
\toprule
regime & observation & practical decision \\
\midrule
small single-turn & tool schema and extra turn can exceed unused mass & use hybrid if all content must remain available; otherwise inline \\
large single-turn & dynamic arms save; static \texttt{original8} is cheaper on SSB & test static reduction first, then hybrid for full coverage \\
small, always-needed multi-turn & cache-correct input edge is small and output-sensitive & treat as near parity; avoid a heavy catalog \\
large, compressible multi-turn & lean on-demand pair reduces repeated footprint sharply & evaluate Skill Block and hybrid under local accounting \\
\bottomrule
\end{tabularx}
\end{center}

This is a \textbf{mechanism-specific post-hoc heuristic}, not a validated gate or a general size cutoff. Hybrid is the small-skill exception because its stub yields a near-zero fetch rate. The decision should use measured demand, context overhead, and provider accounting rather than adopt the approximately 2K pure-Skill-Block threshold from this endpoint. The \texttt{d} sensitivity should be computed before choosing a multi-turn mechanism, and output and latency concerns should be measured locally rather than inferred from input tokens.

\section*{6. Discussion}
Loading and content optimization address different levers. A skill optimizer can improve what each procedure says; conditional loading can avoid presenting irrelevant procedures. The two compose, but neither result establishes that another skill partition, another endpoint, or another prompting convention will show the same ordering. In particular, selection quality is inferred only indirectly from fetch behavior and task outcomes. SpreadsheetBench's execution-feedback loop can repair an imperfect selection, and we do not provide a labeled routing precision/recall evaluation.

The results clarify why the single-turn and multi-turn cases need separate treatment. In a single request, an on-demand mechanism pays its interaction overhead once. Hybrid can be attractive because a stub delivers enough guidance to prevent that interaction; the SearchQA fetch rate is only 1.4\%. By contrast, in stateful interaction, an always-on footprint is repeatedly resubmitted. When a large procedure library is mostly unused, reducing that footprint dominates the occasional request for a needed block. This explanation is \textbf{consistent with} the raw/new/effective logs, but it is not a causal attribution of cache savings to the tool schema: schema and conversational prefix effects cannot be separated in the provider usage data.

The multi-turn results also illustrate why raw usage fields require care. A long resubmitted prefix can make a lean mechanism look expensive under raw tokens even when its newly processed footprint is lower; effective accounting corrects that under an explicit discount sensitivity. Conversely, an accounting correction cannot supply omitted costs: provider billing, latency, separate reasoning tokens, and schema-specific cache attribution remain unmeasured. The practical implication is narrow: record raw, cache-read, new, output, and turn-level data before making cost claims.

The static SpreadsheetBench comparison sharpens the deployment interpretation. A loader is valuable only if optional coverage is actually needed. Historical \texttt{original8} is cheaper on the measured split, but permanently discards six procedures, differs from exact \texttt{CORE\_\allowbreak{}BLOCKS}, and has not been shown to preserve performance under shifted or broader requests. Thus it does not refute the content-parity question; it prevents a stronger claim---that this benchmark demonstrates a need for dynamic routing---that the data do not support.

Finally, the configured-model reruns reproduce large effective-input savings for the on-demand pair on one real and one synthetic benchmark, while their internal ordering changes or is nearly tied. This supports evaluating both mechanisms, not a stable model-specific rank, provider-attested model isolation, or transfer beyond this provider. Extended operational trade-offs and a deployment checklist are in Appendix E.

\section*{7. Limitations \& Threats to Validity}
\begin{enumerate}[leftmargin=1.6em]
\item \textbf{Endpoint and transfer scope.} Primary results use an endpoint reported as GHCP \texttt{gpt-\allowbreak{}5.5}; transfer checks configure \texttt{gpt-\allowbreak{}5.6} on the same provider but do not retain provider-returned model IDs. They are configuration-transfer evidence, not provider-attested model isolation. Cross-vendor replication, repeated seeds, and uncontrolled production drift remain open.
\item \textbf{Production endpoint.} Absolute scores depend on a system prompt and decoding behavior we do not control. Claims are paired within-benchmark comparisons, not public-model reproducibility or leaderboard results.
\item \textbf{Outcome inference.} Non-significant McNemar tests and token CIs do not prove equivalence or non-inferiority. Token intervals are per-comparison rather than simultaneous family-wise intervals. SearchQA and SpreadsheetBench reused final cases during development, making the combined primary outcome family exploratory.
\item \textbf{Synthetic probe and safety confound.} SynthProc measures controlled token mechanics; \texttt{SECRET KEY} refusal loops on \texttt{gpt-\allowbreak{}5.6} make its failures unsuitable as reliability evidence.
\item \textbf{Static control interpretation.} Historical \texttt{original8} is non-content-parity, differs from the dynamic core, is separately timed, and does not demonstrate coverage beyond the measured SpreadsheetBench split; it gives no non-inferiority proof.
\item \textbf{Selection and partitioning.} Block boundaries are author-chosen, and routing correctness is not directly labeled or measured. Task accuracy is an especially weak routing proxy when feedback can recover errors.
\item \textbf{Prompt scaffolding.} Mechanisms require different instructions and catalogs. The study measures loading with its necessary scaffolding; SearchQA demonstrates that guidance can materially change load behavior.
\item \textbf{Accounting assumptions.} \texttt{d=.1} is a reference sensitivity, not internal GHCP billing; schema-cache attribution is unmeasured and described only as consistent with the logs. Output sensitivity omits unobserved historical reasoning tokens and measured latency.
\item \textbf{ScienceWorld execution and repair.} The primary n=84 analysis removes six IDs with arm-specific aborts from a historical run whose failed-call costs and provider-returned model identifiers are incomplete. The n=82 sensitivity additionally removes both IDs containing the two non-aborted episodes recorded above 600 seconds from every arm. The repaired matched n=90 artifact replaces 24 arm-items with later call-audited reruns and is therefore temporally spliced; it is corroborating sensitivity evidence rather than a contemporaneous full rerun.
\item \textbf{Gate status.} The cost gate explains observed signs post hoc and is not a prospectively validated predictor. The hybrid exception precludes a universal size rule.
\end{enumerate}

Appendix E expands these points, including temporal drift, cache writes, and guidance implementation implications.

\section*{8. Conclusion}
For a fixed skill, conditional loading can decouple coverage from per-query context cost, but the right packaging depends on regime. In paired GHCP runs reported as \texttt{gpt-\allowbreak{}5.5}, hybrid is the best content-parity option in two single-turn cells; large compressible multi-turn skills favor the Skill Block/hybrid pair; and small always-needed multi-turn procedures are near parity under total-cost sensitivity. Exploratory static pruning can be cheaper when reduced coverage is acceptable, as SpreadsheetBench \texttt{original8} shows, but it is not a content-parity substitute or an exact static-core control. Caching-correct accounting is essential for multi-turn claims. Results from runs configured as \texttt{gpt-\allowbreak{}5.6} support the large-skill pattern on one synthetic and one real benchmark, while unavailable provider-returned model identity, same-provider scope, and SynthProc's safety confound remain material. The contribution is a bounded regime map and accounting method, not a universal winner.

\section*{References}
\begin{itemize}[leftmargin=1.4em]
\item {}[Agrawal et al. 2025] L. A. Agrawal, S. Tan, D. Soylu, et al. \emph{GEPA: Reflective Prompt Evolution Can Outperform Reinforcement Learning.} arXiv:2507.19457, 2025.
\item {}[Anthropic 2024] Anthropic. \emph{Introducing the Model Context Protocol.} 2024. \url{https://www.anthropic.com/news/model-context-protocol} (spec: \url{https://modelcontextprotocol.io}).
\item {}[Anthropic 2025a] Anthropic. \emph{Introducing Agent Skills.} October 16, 2025. \url{https://claude.com/blog/skills}.
\item {}[Anthropic 2025b] Anthropic. \emph{Equipping agents for the real world with Agent Skills.} October 16, 2025. \url{https://www.anthropic.com/engineering/equipping-agents-for-the-real-world-with-agent-skills}.
\item {}[Côté et al. 2018] M.-A. Côté, Á. Kádár, X. Yuan, et al. \emph{TextWorld: A Learning Environment for Text-based Games.} arXiv:1806.11532, 2018.
\item {}[Dunn et al. 2017] M. Dunn, L. Sagun, M. Higgins, V. Ugur Guney, V. Cirik, K. Cho. \emph{SearchQA: A New Q\&A Dataset Augmented with Context from a Search Engine.} arXiv:1704.05179, 2017.
\item {}[Fedus et al. 2022] W. Fedus, B. Zoph, N. Shazeer. \emph{Switch Transformers: Scaling to Trillion Parameter Models with Simple and Efficient Sparsity.} \emph{Journal of Machine Learning Research}, 23(120):1--39, 2022. arXiv:2101.03961.
\item {}[Jiang et al. 2023] H. Jiang, Q. Wu, C.-Y. Lin, Y. Yang, L. Qiu. \emph{LLMLingua: Compressing Prompts for Accelerated Inference of Large Language Models.} EMNLP 2023. arXiv:2310.05736.
\item {}[Jiang et al. 2024] H. Jiang, Q. Wu, X. Luo, et al. \emph{LongLLMLingua: Accelerating and Enhancing LLMs in Long Context Scenarios via Prompt Compression.} ACL 2024. arXiv:2310.06839.
\item {}[Khattab et al. 2024] O. Khattab, A. Singhvi, P. Maheshwari, et al. \emph{DSPy: Compiling Declarative Language Model Calls into State-of-the-Art Pipelines.} ICLR 2024. \url{https://iclr.cc/virtual/2024/poster/17642}. Preprint: arXiv:2310.03714, titled \emph{DSPy: Compiling Declarative Language Model Calls into Self-Improving Pipelines}.
\item {}[Liu et al. 2024] N. F. Liu, K. Lin, J. Hewitt, et al. \emph{Lost in the Middle: How Language Models Use Long Contexts.} TACL 12, 2024. arXiv:2307.03172.
\item {}[Ma et al. 2024] Z. Ma, B. Zhang, J. Zhang, et al. \emph{SpreadsheetBench: Towards Challenging Real World Spreadsheet Manipulation.} NeurIPS 2024 Datasets \& Benchmarks (spotlight). arXiv:2406.14991.
\item {}[Shazeer et al. 2017] N. Shazeer, A. Mirhoseini, K. Maziarz, et al. \emph{Outrageously Large Neural Networks: The Sparsely-Gated Mixture-of-Experts Layer.} ICLR 2017. arXiv:1701.06538.
\item {}[Shridhar et al. 2020] M. Shridhar, J. Thomason, D. Gordon, et al. \emph{ALFRED: A Benchmark for Interpreting Grounded Instructions for Everyday Tasks.} CVPR 2020. arXiv:1912.01734.
\item {}[Shridhar et al. 2021] M. Shridhar, X. Yuan, M.-A. Côté, Y. Bisk, A. Trischler, M. Hausknecht. \emph{ALFWorld: Aligning Text and Embodied Environments for Interactive Learning.} ICLR 2021. arXiv:2010.03768.
\item {}[Wang et al. 2022] R. Wang, P. Jansen, M.-A. Côté, P. Ammanabrolu. \emph{ScienceWorld: Is your Agent Smarter than a 5th Grader?} EMNLP 2022. arXiv:2203.07540.
\item {}[Yang et al. 2026] Y. Yang, Z. Gong, W. Huang, et al. \emph{SkillOpt: Executive Strategy for Self-Evolving Agent Skills.} arXiv:2605.23904, 2026.
\item {}[Yao et al. 2023] S. Yao, J. Zhao, D. Yu, et al. \emph{ReAct: Synergizing Reasoning and Acting in Language Models.} ICLR 2023. arXiv:2210.03629.
\item {}[Yuksekgonul et al. 2024] M. Yuksekgonul, F. Bianchi, J. Boen, et al. \emph{TextGrad: Automatic "Differentiation" via Text.} arXiv:2406.07496, 2024.
\item {}[Zhou et al. 2023] Y. Zhou, A. I. Muresanu, Z. Han, et al. \emph{Large Language Models Are Human-Level Prompt Engineers.} ICLR 2023. arXiv:2211.01910.
\end{itemize}

\section*{Appendix A: Implementation and Reproducibility}
\subsection*{A.1 Endpoint, artifacts, and protocol}
The regime-map runs use a GHCP endpoint reported as \texttt{gpt-\allowbreak{}5.5} with \texttt{GHCP\_\allowbreak{}LEAN\_\allowbreak{}SYSTEM=1}; historical artifacts do not independently attest all provider-returned model and reasoning settings. The targeted ScienceWorld replacements explicitly configured \texttt{gpt-\allowbreak{}5.5}; the public provenance manifest records that model and \texttt{medium} observed reasoning effort, although the configured effort variable was unset. The transfer runs configure \texttt{GHCP\_\allowbreak{}MODEL=gpt-\allowbreak{}5.6-\allowbreak{}terra} with reasoning effort unset, but do not retain provider-returned model identity; the additional SynthProc robustness configuration sets \texttt{GHCP\_\allowbreak{}REASONING\_\allowbreak{}EFFORT=medium}. The historical SynthProc full, Skill Block, and reference arms completed on July 11 UTC, while the reported hybrid completed July 12 and replaced a distinct \texttt{truehybrid} arm from the earlier parallel sweep. Exact production endpoint internals are unavailable, so reproducible units are paired analysis outputs rather than absolute score calibration.

Commands, file layouts, public harness locations, retained result bundles, and analysis-script usage are documented in the repository README and per-benchmark READMEs. SpreadsheetBench workbook-level correction replay additionally requires retained prediction workbooks and trajectories that are omitted from the public repository for size; the corrected bundle preserves their hashes and comparison evidence.

\subsection*{A.2 Skill format and dependency rendering}
The substrate uses YAML front matter for \texttt{name}, \texttt{description}, \texttt{default}, \texttt{requires}, \texttt{priority}, and \texttt{tags}, and EJS delimiters \texttt{<\%\# block: NAME \%> \ldots{} <\%\# endblock \%>}. A renderer expands default blocks, a requested block, and only prerequisites not already present. Full rendering reconstructs the source content. Skill Block registers a catalog-bearing \texttt{load\_\allowbreak{}skill\_\allowbreak{}block} function; reference exposes a text catalog and injects named \texttt{references/\allowbreak{}} content after a selection turn; hybrid auto-generates one-line stubs from each optional block's description/opening guidance and fetches the complete block through the tool only when required.

The multi-turn substrate also recognizes optional \texttt{steering} and \texttt{execution-\allowbreak{}budget} sections. \texttt{steering} is injected once; \texttt{execution-\allowbreak{}budget} is rendered per turn with \texttt{\{used\}/\allowbreak{}\{max\}/\allowbreak{}\{remaining\}}. Both are ordinary, presence-gated loadable content: a skill without them retains prior behavior. They permit model-specific guidance normalization without changing the loading mechanism, but this paper does not separately ablate that design.

\subsection*{A.3 Per-benchmark packaging}
SearchQA has six blocks: a 20-token overview and five optional blocks; full is approximately 1,966 tokens, Skill Block seeds the overview plus schema, reference uses core plus a text catalog, and hybrid uses an approximately 385-token stub. SpreadsheetBench has 14 blocks: exact dynamic \texttt{CORE\_\allowbreak{}BLOCKS} (approximately 3,513 tokenizer-proxy tokens) plus six specialized blocks; full is approximately 8,116. Dynamic catalogs exclude the seeded core; reference begins with an approximately 526-token overview-only selection context; hybrid uses an approximately 4,029-token stub. Historical static \texttt{original8} is approximately 3,195 tokens and uses a different eight-block set.

ALFWorld has an overview and six $\approx$99-token procedure walkthroughs; full is approximately 1,087 tokens. ScienceWorld has an overview and 16 $\approx$289-token procedures; full is approximately 6,097 tokens. In both, reference includes one-line hints and reads full walkthroughs after selection, while retained \texttt{truehybrid} exposes a selected hint before an optional full reference. SynthProc has an overview and 45 operation blocks, each carrying a hint and key detail; explicit \texttt{o200k\_base} counts are approximately 9,728 full, 6,365 reference catalog, and 1,297 hybrid stub tokens. The historical harness also emitted nearby four-characters-per-token estimates. These proxy sizes can differ from provider usage.

\section*{Appendix B: Full \texttt{gpt-\allowbreak{}5.5} Benchmark Tables and Paired Intervals}
\subsection*{B.1 Single-turn results (raw input)}
\begin{center}\small
\begin{tabularx}{\linewidth}{LrrrrL}
\toprule
benchmark / arm & primary outcome & secondary outcome & input & output & $\Delta$ vs full (95\% paired CI) \\
\midrule
SearchQA hybrid & .845 & .912 & 4,016 & 307 & $-$27.4\% [$-$1,556, $-$1,479] \\
SearchQA full & .841 & .907 & 5,535 & 380 & --- \\
SearchQA reference & .849 & .913 & 6,112 & 628 & +10.4\% [+552, +601] \\
SearchQA Skill Block & .841 & .906 & 8,216 & 940 & +48.4\% [+2,534, +2,827] \\
SSB static \texttt{original8} & .811 & .903 & 5,547 & 1,946 & $-$55.5\% [$-$7,556, $-$6,285] \\
SSB hybrid & .807 & .909 & 7,506 & 1,858 & $-$39.8\% [$-$5,574, $-$4,344] \\
SSB Skill Block & .811 & .905 & 8,028 & 1,967 & $-$35.6\% [$-$5,107, $-$3,773] \\
SSB reference & .807 & .902 & 8,517 & 2,210 & $-$31.7\% [$-$4,553, $-$3,341] \\
SSB full & .800 & .896 & 12,461 & 2,292 & --- \\
\bottomrule
\end{tabularx}
\end{center}

Percentages are aggregate deltas versus full; bracketed intervals are paired token-difference CIs.
For SearchQA the outcome columns are hard/soft answer match. For SpreadsheetBench they are hard workbook pass/cell accuracy on n=275; token and output means are on the separate n=276 token-complete set.

SearchQA hybrid is cheaper on 1,380/1,400 cases. Skill Block loads at least one block on 75\% of cases and averages 2.05 fetches; hybrid averages .02 and loads any block on 1.4\%. Corrected Skill Block versus full McNemar discordances are 25/26 (exact p=1.0). On the SSB token-complete set, specialized-block selection occurs in 97/276 Skill Block, 47/276 hybrid, and 254/276 reference cases. Static \texttt{original8} is cheaper than hybrid on 236/276 cases. On the 275 outcome-clean cases, its McNemar p-value is 1.0 versus each dynamic arm and .678 versus full; temporal drift and treatment mismatch remain caveats.

\subsection*{B.2 Multi-turn results (median raw / new / effective)}
\begin{center}\small
\begin{tabularx}{\linewidth}{LrrrrrrL}
\toprule
benchmark / arm & success & progress & raw & new & effective & output & aggregate $\Delta$; paired difference (95\% CI) \\
\midrule
ALF full & 42/42 & --- & 28,900 & 3,901 & 6,368 & 534 & --- \\
ALF Skill Block & 42/42 & --- & 32,372 & 2,639 & 5,569 & 631 & $-$12.55\% [$-$15.93\%, $-$7.45\%]; $-$907 [$-$992, $-$876] \\
ALF hybrid & 42/42 & --- & 34,649 & 2,980 & 6,166 & 778 & $-$3.18\% [$-$9.50\%, +1.29\%]; $-$336 [$-$455, $-$208] \\
ALF reference & 42/42 & --- & 37,754 & 4,562 & 7,907 & 626 & +24.17\% [+19.41\%, +31.05\%]; +1,188 [+965, +1,306] \\
Science full & 57/84 & .901 & 149,466 & 12,892 & 26,810 & 1,111 & --- \\
Science Skill Block & 62/84 & .916 & 59,197 & 4,575 & 10,061 & 1,438 & $-$62.5\%; $-$15,322 [$-$16,488, $-$13,603] \\
Science hybrid & 58/84 & .902 & 72,181 & 5,780 & 12,654 & 2,312 & $-$52.8\%; $-$13,252 [$-$14,435, $-$11,813] \\
Science reference & 50/84 & .833 & 127,248 & 8,974 & 20,644 & 2,276 & $-$23.0\%; $-$9,833 [$-$11,245, $-$8,607] \\
Synth full & 40/40 & --- & 65,380 & 16,754 & 21,617 & 536 & --- \\
Synth Skill Block & 40/40 & --- & 29,912 & 3,006 & 5,837 & 1,347 & $-$73.0\%; $-$15,871 [$-$16,398, $-$15,264] \\
Synth hybrid & 40/40 & --- & 28,020 & 4,686 & 7,221 & 1,344 & $-$66.6\%; $-$14,384 [$-$15,244, $-$13,971] \\
Synth reference & 40/40 & --- & 70,528 & 12,642 & 18,428 & 702 & $-$14.8\%; $-$3,191 [$-$4,014, $-$1,681] \\
\bottomrule
\end{tabularx}
\end{center}

Percentages are delta-of-aggregate medians versus full; the following token quantity and bracketed interval are the separately labeled paired-median-difference estimand. For primary ScienceWorld, the estimand-aligned percentage CIs are [$-$67.76\%, $-$57.44\%], [$-$64.74\%, $-$37.66\%], and [$-$40.21\%, +4.49\%] for Skill Block, hybrid, and reference.

ALFWorld has paired median deltas of $-$907 Skill Block, $-$336 hybrid, and +1,188 reference; the corresponding arms are cheaper on 38/42, 34/42, and 5/42 episodes. Primary ScienceWorld arms are cheaper on 81/84, 73/84, and 69/84. The strict n=82 sensitivity gives Skill Block/hybrid/reference reductions of 62.05\%/55.85\%/25.57\%, with reference again including zero. The repaired matched n=90 sensitivity is 62.85\%/50.84\%/25.38\% cheaper, with 87/90, 79/90, and 75/90 cheaper episodes, respectively. SynthProc's Skill Block--hybrid effective delta is $-$1,328 (CI [$-$1,460, $-$1,221]) and Skill Block is cheaper on 37/40.

\section*{Appendix C: Accounting and Robustness Details}
\subsection*{C.1 Cache fractions, means, and output sensitivity}
\begin{center}\small
\begin{tabularx}{\linewidth}{LLrrr}
\toprule
benchmark & cache-read fraction & SB mean $\Delta$ & hybrid mean $\Delta$ & ref mean $\Delta$ \\
\midrule
ALFWorld & .87--.92 & $-$13.6\% & $-$2.4\% & +18.5\% \\
ScienceWorld & .93--.94 & $-$53.3\% & $-$37.7\% & $-$20.2\% \\
SynthProc & .74--.90 & $-$73.1\% & $-$67.0\% & $-$14.9\% \\
\bottomrule
\end{tabularx}
\end{center}

\begin{center}\small
\begin{tabularx}{\linewidth}{Lrrr}
\toprule
benchmark & Skill Block total\_$\lambda$ & hybrid total\_$\lambda$ & reference total\_$\lambda$ \\
\midrule
SearchQA $\lambda$=4 & +69.7\% & $-$25.7\% & +22.2\% \\
SSB $\lambda$=4 & $-$26.5\% & $-$30.9\% & $-$19.8\% \\
ALF $\lambda$=4 / $\lambda$=8 & $-$4.5\% / $-$0.6\% & +9.2\% / +15.2\% & +21.3\% / +20.4\% \\
Science $\lambda$=4 / $\lambda$=8 & $-$46.8\% / $-$37.2\% & $-$24.5\% / $-$8.3\% & $-$2.7\% / +12.5\% \\
Synth $\lambda$=4 / $\lambda$=8 & $-$53.7\% / $-$37.2\% & $-$48.0\% / $-$31.0\% & $-$10.2\% / $-$6.8\% \\
\bottomrule
\end{tabularx}
\end{center}

$total_\lambda $ applies raw input in the single-turn rows and effective input in multi-turn rows. $\lambda$=1 is also recomputed in the analysis artifact; $\lambda$=4 and $\lambda$=8 are shown because they bound the main interpretation. Output and hidden reasoning accounting are sensitivity analyses, not GHCP invoices.

\subsection*{C.2 Discount and cache-write sensitivity}
At \texttt{d=1} raw input is the upper-bound input cost. Raw input remains the methodological primary metric for the single-turn benchmarks. If the paper's cache-discount formula is nevertheless applied mechanically to SearchQA, Skill Block and reference cross the full arm near $d=.354$ and $d=.440$, respectively; their cost ordering is therefore cache-policy sensitive. ScienceWorld and SynthProc Skill Block beat full on raw input (ScienceWorld approximately $-$60\%; SynthProc approximately $-$54\%), so their signs persist from the reference discount through raw accounting. ALFWorld is discount-sensitive: the marginal-median crossovers are $d=.268$ for Skill Block and $.133$ for hybrid, while paired-difference crossovers are $.327$ and $.151$. No cache-write premium is added; newly processed tokens are charged at full rate. ALF median write counts are full 3,884, Skill Block 2,619, hybrid 2,960, and reference 4,543.5. A hypothetical write premium would not reduce the observed lead of the two smaller-write arms, but provider write pricing is not measured.

\subsection*{C.3 Statistical details}
Bootstrap intervals in Appendix B are 95\% paired-resampling intervals and are not simultaneous family-wise token intervals. Single-turn paired mean differences use the corrected per-benchmark machine-readable analyses and fixed seeds recorded with each bundle; \texttt{analyze\_\allowbreak{}paired\_\allowbreak{}singleturn.\allowbreak{}py} reproduces the consolidated table. Multi-turn reporting distinguishes two estimands: the displayed difference of marginal medians is bootstrapped directly, while median paired differences and their intervals are reported separately. Both use fixed-seed 100,000-resample percentile bootstraps. The nine principal gpt-5.5 outcome comparisons across SearchQA, SpreadsheetBench, and primary common non-aborted ScienceWorld use exact paired McNemar tests with Holm adjustment; all adjusted p-values are 1.0. Because final benchmark cases informed SearchQA/SpreadsheetBench development and SpreadsheetBench block selection, these inferential results are exploratory. Configured-\texttt{gpt-\allowbreak{}5.6} ScienceWorld's three outcome comparisons are a separate Holm family, each adjusted p=.116. These results warrant ``no detected difference'' only; they do not supply a pre-registered non-inferiority margin.

\section*{Appendix D: Configured-Model Full Details}
\subsection*{D.1 SynthProc configured as \texttt{gpt-\allowbreak{}5.6}}
\begin{center}\small
\begin{tabularx}{\linewidth}{Lrrrrrr}
\toprule
arm & success & raw & effective & output & reasoning & $\Delta$ effective vs full \\
\midrule
full & 37/40 & 42,454 & 14,053 & 249 & 106 & --- \\
Skill Block & 40/40 & 16,841 & 4,725 & 296 & 93 & $-$66.4\% \\
hybrid & 40/40 & 15,558 & 4,731 & 270 & 72 & $-$66.3\% \\
reference & 39/40 & 45,286 & 11,760 & 234 & 66 & $-$16.3\% \\
\bottomrule
\end{tabularx}
\end{center}

At $\lambda$=4, Skill Block and hybrid are $-$60.9\% and $-$60.4\%; at $\lambda$=8 they are $-$56.1\% and $-$57.2\%. The medium-effort supplementary run gives hybrid $-$68.4\%, Skill Block $-$66.4\%, and reference $-$8.9\%. The default run is paired within configuration, but provider-returned model identity was not retained. The key-specific classifier affects 13 default full episodes (10 successful), 2 default reference episodes (1 successful), 11 medium full episodes (6 successful), and 4 medium reference episodes (3 successful). It flags no Skill Block/hybrid episode, but one default hybrid episode contains generic inability language. Thus the wording-sensitive confound affects measured token cost in addition to all observed failures; no reliability or clean loading-effect conclusion should be drawn from it.

Benchmark integrity requires evaluator isolation: the model receives only its arm-specific rendered skill, registered tool results, and current observation. It must not have filesystem access to \texttt{ops.\allowbreak{}json}, \texttt{tasks.\allowbreak{}jsonl}, \texttt{tiers.\allowbreak{}json}, the unrestricted full skill, or historical result trajectories, all of which contain synthetic evaluator ground truth.

Paired median effective deltas are $-$10,013 for Skill Block (CI [$-$10,456, $-$9,335], cheaper 40/40), $-$10,329 for hybrid (CI [$-$11,217, $-$9,639], 40/40), and $-$3,279 for reference (CI [$-$4,303, $-$2,278], 35/40).

\subsection*{D.2 Real ScienceWorld configured as \texttt{gpt-\allowbreak{}5.6}}
\begin{center}\small
\begin{tabularx}{\linewidth}{Lrrrrrr}
\toprule
arm & success & progress & raw & effective & output & $\Delta$ effective vs full \\
\midrule
full & 52/90 & .859 & 128,994 & 28,459 & 656 & --- \\
hybrid & 60/90 & .904 & 40,198 & 8,869 & 604 & $-$68.8\% \\
Skill Block & 58/90 & .875 & 43,382 & 9,173 & 600 & $-$67.8\% \\
reference & 60/90 & .896 & 58,805 & 11,812 & 574 & $-$58.5\% \\
\bottomrule
\end{tabularx}
\end{center}

Skill Block's paired median delta is $-$13,916, CI [$-$16,014, $-$11,900], cheaper on 89/90; hybrid's is $-$13,683, CI [$-$16,645, $-$11,020], cheaper on 89/90; reference's is $-$11,461, CI [$-$13,241, $-$8,071], cheaper on 88/90. Mean effective deltas are $-$58.5\%, $-$61.1\%, and $-$49.1\% for Skill Block, hybrid, and reference. At $\lambda$=4 they are $-$63.9\%, $-$63.4\%, and $-$54.6\%; at $\lambda$=8, $-$59.8\%, $-$58.9\%, and $-$50.9\%. Outcome tests remain non-significant after Holm adjustment (p=.116). This is real-environment, same-provider evidence for the large-skill pattern, not cross-vendor evidence.

\section*{Appendix E: Extended Limitations and Design Details}
\textbf{Prompt scaffolding and guidance.} Different mechanisms must instruct the agent to call a tool, request a reference, or use a stub; these minimal changes are part of deployment, not nuisance text that can be removed. The frozen aggressive SearchQA loading nudge yields +48.4\% for pure Skill Block. Retained first-200 pilots used alternative guidance and loaded less often, but they are neither controlled full-test ablations nor confirmatory evidence for a conservative nudge. The reported aggressive run is an overhead diagnostic, not a recommended configuration. Steering and execution-budget sections are loadable content, but their placement has not been experimentally isolated.

\textbf{ALFWorld sampling and action execution.} The 42 episodes are the first seven lexicographically sorted games from each of six task types; seed 42 does not determine membership, so conclusions are scoped to this deterministic balanced head subset. The harness maps free-form output to admissible commands. Three of 1,512 actions change, all the same semantic correction from \texttt{go to shelf 3} to \texttt{go to shelf 9} in one item across full, Skill Block, and hybrid; historical records preserve raw and executed text but not a normalization-reason field.

\textbf{Schema cache attribution.} The observed raw/new/effective patterns are consistent with provider prefix caching of repeated context, including repeated tool definitions, but the logs do not attribute cache hits to schema versus conversation content. Providers that bill schemas or cache writes differently can move the boundary between Skill Block and hybrid. We therefore use ``consistent with'' rather than a causal schema-cache claim.

\textbf{Size gate and post-hoc model.} The approximately 2K pure-Skill-Block heuristic reflects observed schema and round-trip overhead on this endpoint. It does not apply to hybrid, whose stub can reduce fetches to nearly zero, and it has not been swept as a break-even curve. The model organizes outcomes using observed load behavior; it is not a preregistered prediction.

\textbf{Temporal and endpoint drift.} Several multi-turn arm processes were launched concurrently, but calls were not randomized or interleaved at the item level; the historical SynthProc hybrid and static SpreadsheetBench control ran later. Their large token gaps are unlikely to be explained entirely by drift, while smaller outcome differences cannot be separated cleanly from it. Endpoint changes, model updates, and one-run-per-configured-model transfer checks are reasons to avoid stable-rank claims between near-tied Skill Block and hybrid.

\textbf{Other unresolved scope.} Blocks are mostly hand authored (only SpreadsheetBench uses a SkillOpt-derived skill), and their boundaries determine optional mass. Semantic selection is intended to handle natural-language variation, but multilingual and paraphrase routing robustness were not directly tested. We do not measure wall-clock latency, provider invoices, or a separate historical reasoning-token field. Cross-vendor replication, learned block partitioning, an oracle/random routing control, and a labeled natural-task routing evaluation remain the clearest next experiments.

\textbf{Mechanism failure modes.} Full loading avoids selection failure but pays for all coverage. Skill Block adds a structured tool decision; reference removes the function schema but retains a textual catalog and selection interaction; hybrid pays a recurring stub cost and can still fetch details. The relevant treatment includes this interaction protocol.

\textbf{Availability versus necessity.} The four primary SpreadsheetBench arms preserve all 14 blocks. Historical static \texttt{original8} removes six blocks, but is not the exact dynamic core; it cannot establish coverage under shift, equivalence, or the causal effect of removing only retrieval. Dynamic fetches show that selection occurred, not that every fetched block was causally necessary.

\textbf{Benchmark complementarity.} ScienceWorld supplies real variable-outcome evidence; SynthProc isolates per-step information demand. Agreement supports the large-skill regime, while the synthetic safety confound prevents reliability interpretation.

\textbf{Synthetic evaluator isolation.} SynthProc ground-truth files and historical trajectories contain synthetic operation/code mappings. Exposing them to the evaluated model would be benchmark-answer leakage; reported runs restrict the model to arm-specific context and environment observations.

\textbf{Deployment checklist.} Estimate core size, optional-demand frequency, expected fetch turns, task-context size, and repeated-prefix behavior; test static reduction if coverage permits; then compare content-parity mechanisms under local cache, output, and latency accounting.

\end{document}